\pdfoutput=1

\documentclass[11pt]{article}

\usepackage[final]{acl}

\usepackage{times}
\usepackage{latexsym}
\usepackage{float}
\usepackage[T1]{fontenc}

\usepackage[utf8]{inputenc}

\usepackage{microtype}
\usepackage{subcaption}
\usepackage{inconsolata}

\usepackage{graphicx}
\usepackage[nameinlink]{cleveref}
\usepackage{xurl}
\usepackage{xspace}

\newcommand{\system}{\textsc{WildVis}\xspace}

\title{\system: Open Source Visualizer for Million-Scale Chat Logs in the Wild}

\author{%
 Yuntian Deng$^{1*}$, Wenting Zhao$^2$, Jack Hessel$^3$, \\\textbf{Xiang Ren$^{4}$, Claire Cardie$^2$, Yejin Choi$^{5,6*}$} \\[1ex]
$^{1}$University of Waterloo \quad $^{2}$Cornell University \quad $^{3}$Samaya AI \\ $^{4}$University of Southern California \quad $^{5}$University of Washington\quad $^{6}$Nvidia\\[1ex]
\small{\texttt{yuntian@uwaterloo.ca, wzhao@cs.cornell.edu, jmhessel@gmail.com}}\\
\small{\texttt{xiangren@usc.edu, cardie@cs.cornell.edu, yejin@cs.washington.edu}}
}


\begin{document}
\maketitle

\def\thefootnote{*}\footnotetext{Work done in large part while at the Allen Institute for Artificial Intelligence.}\def\thefootnote{\arabic{footnote}}

\begin{abstract}
The increasing availability of real-world conversation data offers exciting opportunities for researchers to study user-chatbot interactions. However, the sheer volume of this data makes manually examining individual conversations impractical. To overcome this challenge, we introduce \system, an interactive tool that enables fast, versatile, and large-scale conversation analysis.
\system provides search and visualization capabilities in the text and embedding spaces based on a list of criteria. To manage million-scale datasets, we implemented optimizations including search index construction, embedding precomputation and compression, and caching to ensure responsive user interactions within seconds. We demonstrate \system' utility through three case studies: facilitating chatbot misuse research, visualizing and comparing topic distributions across datasets, and characterizing user-specific conversation patterns.
\system is open-source and designed to be extendable, supporting additional datasets and customized search and visualization functionalities.

\end{abstract}

\section{Introduction}
While hundreds of millions of users interact with chatbots like ChatGPT~\citep{malik2023chatgpt}, the conversation logs remain largely opaque for open research, limiting our understanding of user behavior and system performance.
Recently, initiatives such as WildChat~\citep{zhao2024wildchat} and LMSYS-Chat-1M~\citep{zheng2024lmsyschatm} have released millions of real-world user-chatbot interactions, offering rich opportunities to study interaction dynamics.
However, the volume and complexity of these datasets pose significant challenges for effective analysis.


\begin{figure}[t]
    \centering
    \includegraphics[width=0.7\linewidth]{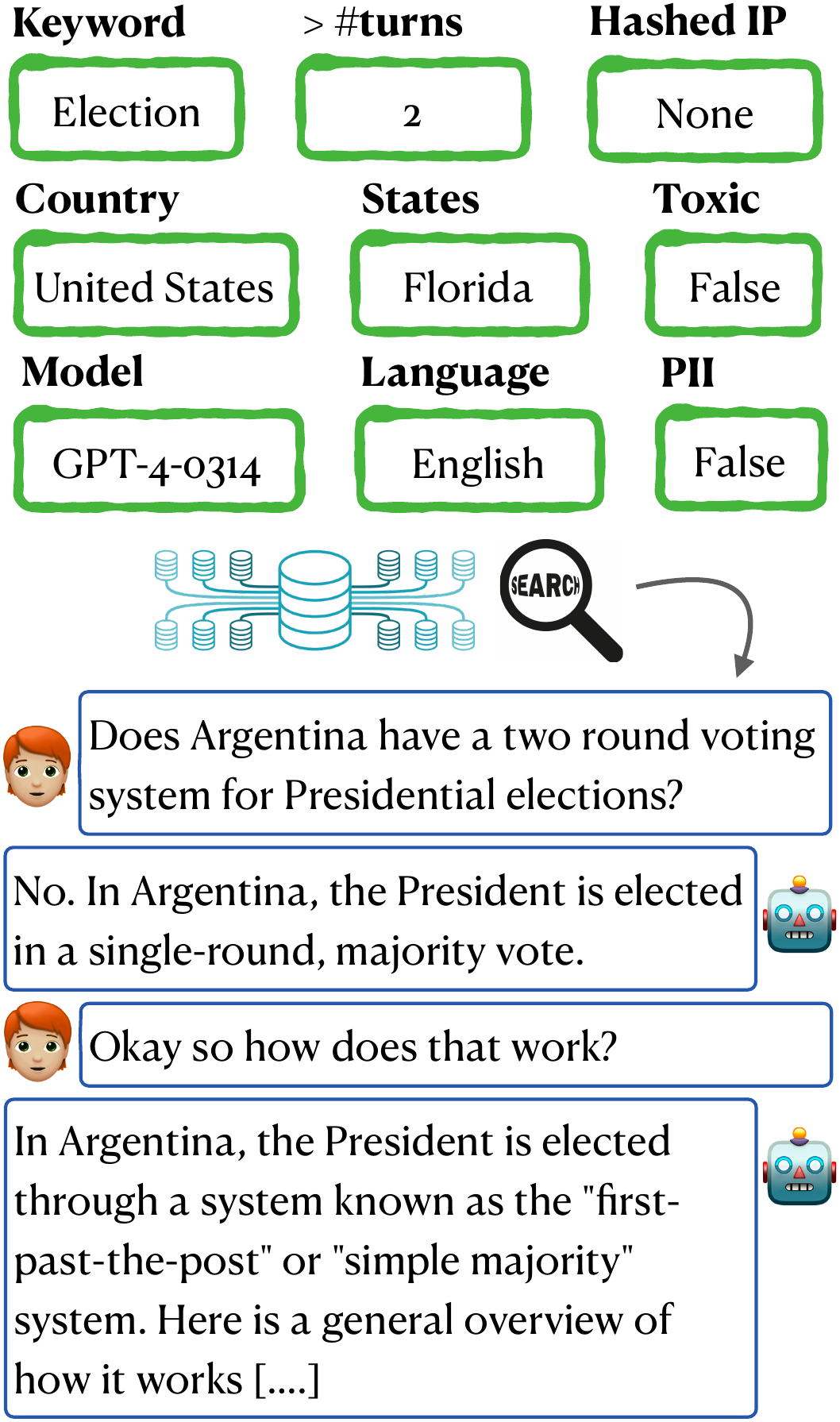}
    \caption{\label{fig:overview}Illustration of an exact, compositional filter-based search in \system. This example demonstrates the application of multiple criteria, including the keyword ``Election,'' conversations with more than two turns, and chats from users in Florida.}   
\end{figure}


To help researchers uncover patterns and anomalies within these vast chat datasets, we introduce \system, an interactive tool for exploring million-scale chat logs. \system enables researchers to find conversations based on specific criteria, understand topic distributions, and explore semantically similar conversations, all while maintaining efficiency. \Cref{fig:overview} illustrates an example search using \system, applying criteria such as the keyword ``Election,'' conversations with more than two turns, and chats from users in Florida, among others.


\begin{figure*}[!t]
    \centering
    \includegraphics[width=\linewidth]{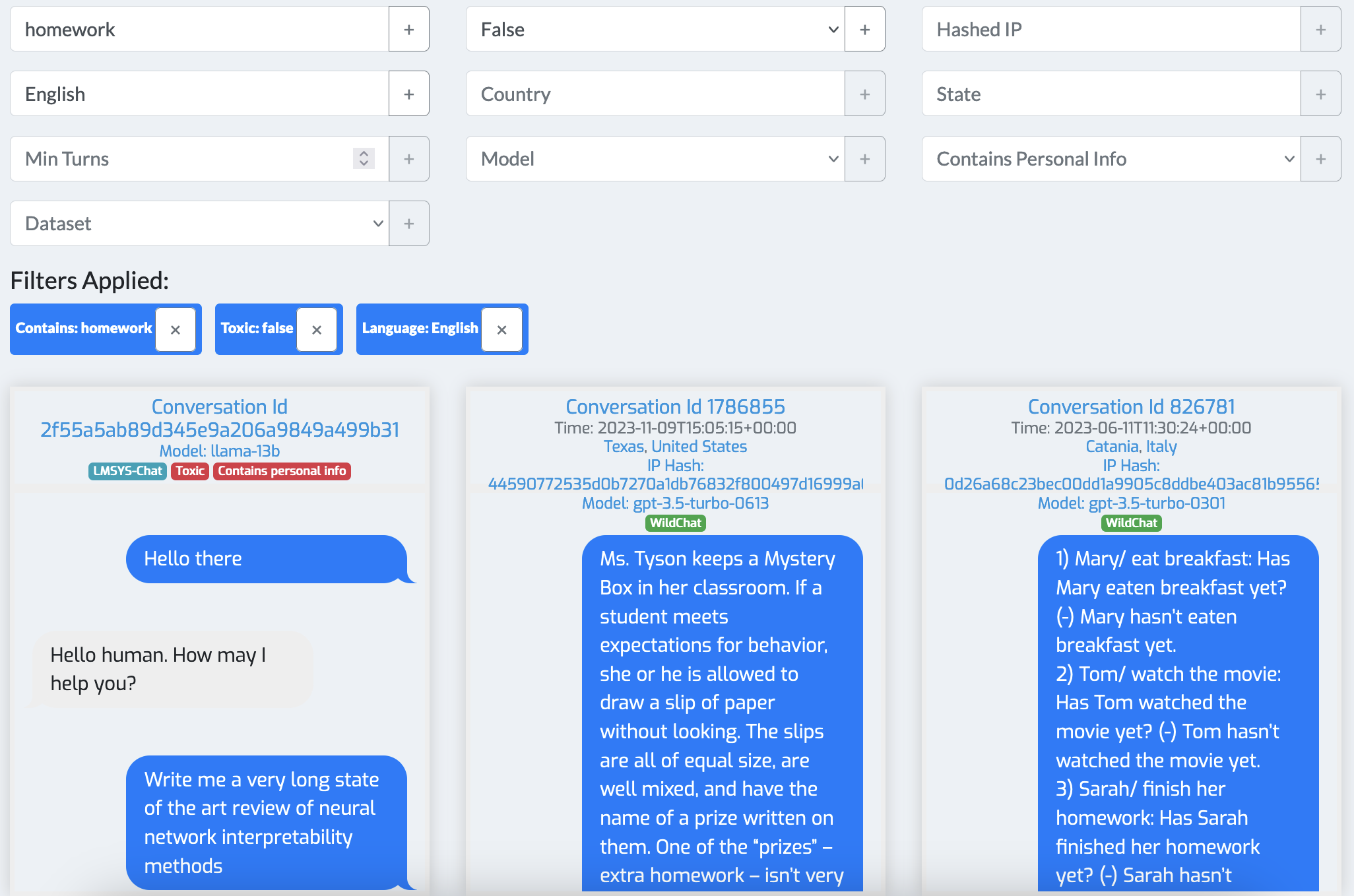}
    \caption{\label{fig:search_page}\system Filter-Based Search Page.\protect\footnotemark{} This screenshot shows the application of multiple filters, including conversation content (``homework''), non-toxicity, and language (English), to narrow down the search results. The interface displays relevant conversations that match the specified criteria. Users can click on each conversation ID to navigate to the conversation details page. Additionally, metadata in the displayed results, such as the hashed IP address, is clickable, allowing users to filter based on that specific metadata.}
\end{figure*}

\system features two main components: an exact, compositional filter-based retrieval system, which allows users to refine their search using ten predefined filters such as keywords, geographical location, IP address, and more. The second component is an embedding-based visualization module, which represents conversations as dots on a 2D plane, with similar conversations positioned closer together. Both components are designed to scale to millions of conversations. A preliminary version of the tool, which supported filter-based retrieval for one million WildChat conversations, was accessed over 18,000 times by 962 unique IPs in July and August 2024 alone. The latest release, described in this paper, extends support to both components for WildChat and LMSYS-Chat-1M.


In this paper, we present the design and implementation of \system, discussing the strategies employed to scale to million-scale datasets while maintaining latency within seconds. We also showcase several use cases: facilitating chatbot misuse research \citep{brigham2024breakingnewscasestudies,mireshghallah2024trustbotdiscoveringpersonal}, visualizing and comparing topic distributions between WildChat and LMSYS-Chat-1M, and characterizing user-specific conversation patterns. For example, \system reveals distinct topic clusters such as Midjourney prompt generation in WildChat and chemistry-related conversations in LMSYS-Chat-1M. Additionally, we observe that WildChat exhibits a generally more creative writing style compared to LMSYS-Chat-1M. As an open-source project, \system is available at \href{https://github.com/da03/WildVisualizer}{github.com/da03/WildVisualizer} under an MIT license, and a working demo can be accessed at \href{https://wildvisualizer.com}{wildvisualizer.com}.

\section{User Interface}
\system consists of two primary pages---a filter-based search page and an embedding visualization page---along with a conversation details page. These pages are designed to provide users with both high-level overviews and detailed insights into individual conversations. 

\footnotetext{This example is available at \url{https://wildvisualizer.com/?contains=homework&toxic=false&language=English}.}

\subsection{Filter-Based Search Page}

\begin{figure*}[!t]
    \centering
    \includegraphics[width=\linewidth]{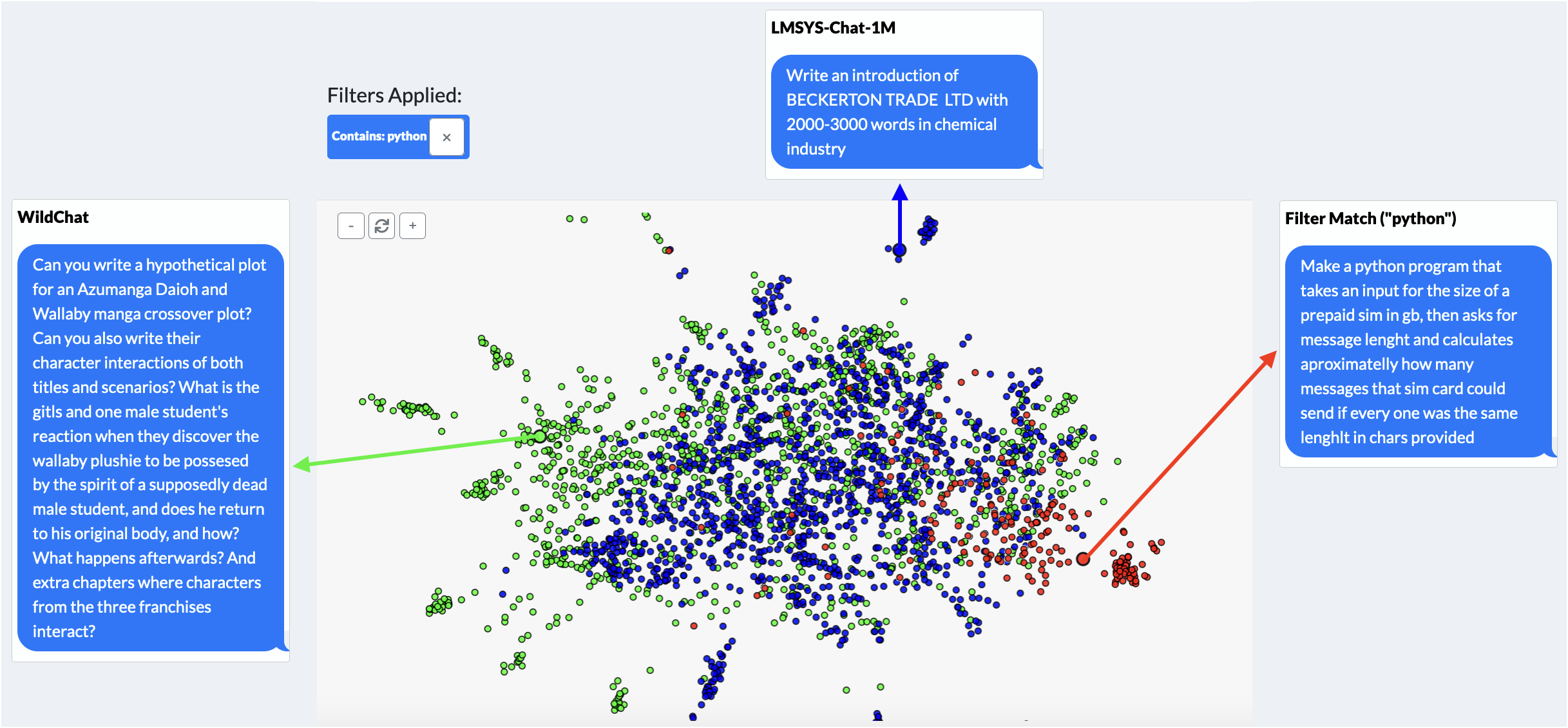}
    \caption{\label{fig:embedding_page}\system Embedding Visualization page.\protect\footnotemark{} Each dot represents a conversation, with green dots from WildChat, blue dots from LMSYS-Chat-1M, and red dots highlighting conversations that match the applied filters (containing ``python'' in this example). Users can interact with the visualization by hovering over dots to preview a conversation and clicking on a dot to navigate to the full conversation. This figure has been enhanced to show a representative example from each category: ``WildChat,'' ``LMSYS-Chat-1M,'' and ``Filter Match.''}
\end{figure*}

The filter-based search page (\Cref{fig:search_page}) enables users to filter the dataset based on a list of criteria. Users can input keywords to retrieve relevant conversations or narrow down results using specific criteria. In total, ten predefined filters are available, including:

\footnotetext{This example is available at \url{https://wildvisualizer.com/embeddings/english?contains=python}.}

\begin{itemize}
    \item Hashed IP Address: Filter conversations by hashed IP addresses to analyze interactions from the same user.\footnote{IP addresses are hashed to protect user privacy while still allowing the analysis of interactions associated with the same user.}
    \item Geographical Data: Filter by inferred state and country to gain insights into regional variations in conversational patterns.
    \item Language: Restrict results to conversations in specific languages.
    \item Toxicity: Include or exclude conversations flagged as toxic.
    \item Redaction Status: Include or exclude conversations with redacted personally identifiable information (PII).
    \item Minimum Number of Turns: Focus on conversations with a specified minimum number of turns.
    \item Model Type: Select conversations by the underlying language model used, such as GPT-3.5 or GPT-4. 
\end{itemize}

The search results are displayed in a paginated table format, ensuring easy navigation through large datasets. Active filters are prominently displayed above the results and can be removed by clicking the ``$\times$'' icon next to each filter.

Each result entry displays key metadata, including the conversation ID, timestamp, geographic location, hashed IP address, and model type. Users can interact with these results in multiple ways. Clicking on a conversation ID leads to a detailed view of that conversation. Additionally, all metadata fields, such as the hashed IP address, are clickable, enabling users to quickly search based on specific attributes. For example, clicking on a hashed IP address brings up a list of all conversations associated with that IP, facilitating user-specific analyses.


\subsection{Embedding Visualization Page}
In addition to traditional search capabilities, \system offers an embedding visualization page (\Cref{fig:embedding_page}), which allows users to explore conversations based on their semantic similarity. Conversations are represented as dots on a 2D plane, with similar conversations placed closer together.

\begin{figure*}[!t]
    \centering
    \includegraphics[width=0.9\linewidth]{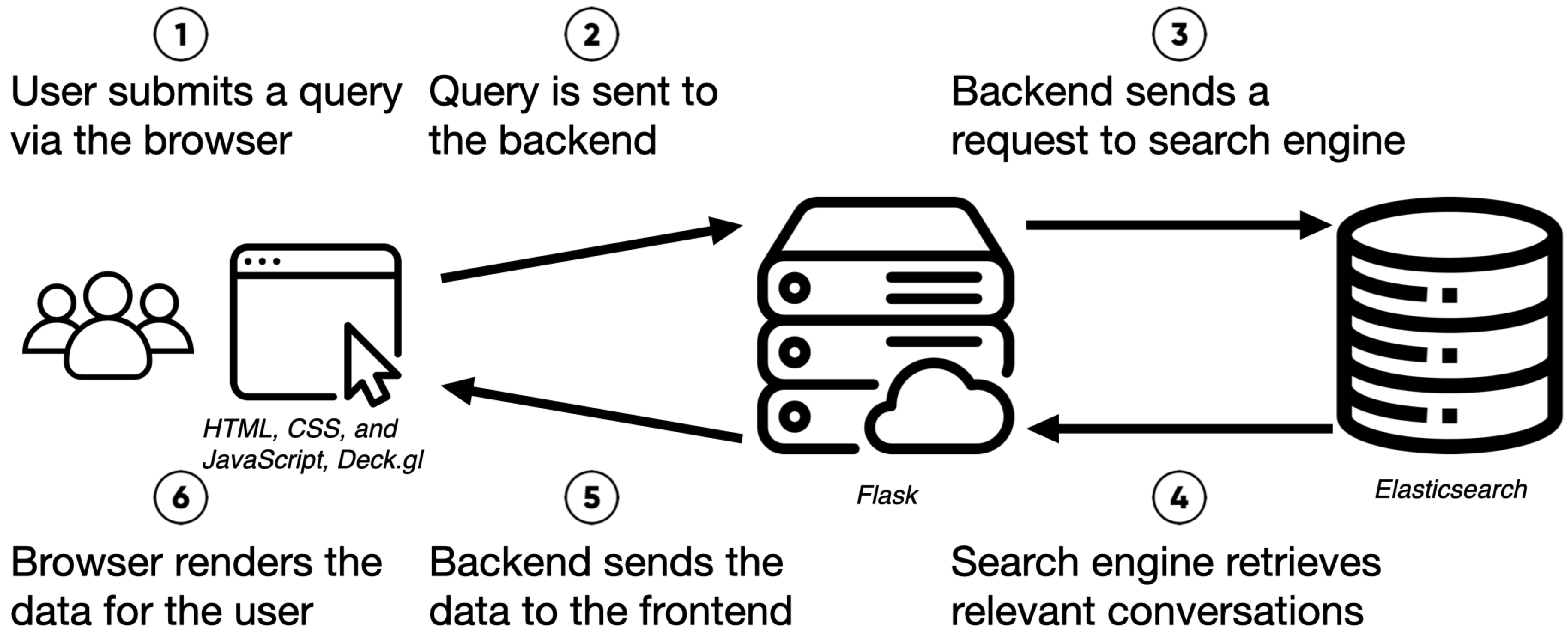}
    \caption{\label{fig:system}System Architecture: Overview of the data flow from user query submission to result rendering in the browser. The software tools used in the frontend, backend, and search engine are italicized.} 
\end{figure*}

\paragraph{Basic Visualization} Each conversation appears as a dot, with different datasets distinguished by color. Hovering over a dot reveals a preview of the conversation, and clicking on it navigates to the conversation details page.\footnote{On mobile devices, tapping a dot displays a preview with options to view the full conversation or close the preview. See \Cref{fig:embedding_page_mobile} in \Cref{appendix:embedding_page_mobile} for a screenshot.} Users can zoom in, zoom out, and drag the view to explore different regions of the visualization. This spatial arrangement enables users to explore clusters of related conversations and identify structures within the data.

\paragraph{Filter-Based Highlighting} Similar to the filter-based search page, users can apply filters to highlight specific conversations on the 2D map, with matching conversations marked in red. This feature helps users locate conversations of interest, such as identifying topics associated with a particular user.

\paragraph{Conversation Embedding} To represent each conversation as a point in 2D space, we embed the first user turn of each conversation using OpenAI's text-embedding-3-small model.\footnote{We opted to embed only the first user turn, as preliminary experiments showed that embedding the entire conversation led to less intuitive clustering.} We then trained a parametric UMAP model~\citep{sainburg2021parametricumapembeddingsrepresentation,mcinnes2020umapuniformmanifoldapproximation} to project these embeddings into 2D space.\footnote{We chose parametric UMAP over t-SNE~\citep{JMLR:v9:vandermaaten08a} to enable online dimensionality reduction, which will be discussed in~\Cref{sec:scalability}.} Since initial experiments showed that training a single UMAP model on all embeddings resulted in some clusters driven by language differences (see \Cref{fig:embedding_language_clusters} in \Cref{appendix:embedding_language}), in order to create more semantically meaningful clusters, we also trained a separate parametric UMAP model for each language. Users can easily switch between different languages and their corresponding UMAP projections (\Cref{fig:embedding_switch} in \Cref{appendix:embedding_switch}).

The combination of embedding visualization, filtering, highlighting, and interactive previews enables users to navigate vast amounts of conversation data, uncovering insights and connections that might otherwise remain hidden. For example, users can easily identify outliers and clusters.



\subsection{Conversation Details Page}
The conversation details page (\Cref{fig:conversation_page} in \Cref{sec:appendix_A}) provides a detailed view of individual conversations. This page displays all the turns between the user and the chatbot, along with associated metadata. Similar to the filter-based search page, all metadata fields are clickable, allowing users to apply filters based on their values. However, if users arrive at this page by clicking a dot on the embedding visualization page, the filtering will be applied within the embedding visualization context. A toggle switch on the conversation details page allows users to control which page (filter-based search or embedding visualization) clicking on metadata fields will direct them to.




\section{System Implementation}
\system is designed to efficiently process large-scale conversational datasets.

\subsection{System Architecture}
\system operates on a client-server architecture, where the server handles data processing, search, and conversation embedding, while the client provides an interface for data exploration. The high-level system architecture is illustrated in \Cref{fig:system}.

Users interact with the frontend web interface, which communicates their queries to the backend server. The backend server is built using Flask\footnote{\url{https://flask.palletsprojects.com/}}, which processes these queries and constructs search requests for an Elasticsearch\footnote{\url{https://www.elastic.co/elasticsearch}} engine. Elasticsearch, known for its scalable search capabilities, retrieves the relevant conversations, which are then sent back to the frontend for rendering. The frontend is developed using HTML, CSS, and JavaScript\footnote{The frontend is built on top of MiniConf~\citep{rush2020miniconfvirtualconference}.}, with Deck.gl\footnote{\url{https://deck.gl/}} used for rendering large-scale, interactive embedding visualizations.

\subsection{\label{sec:scalability}Scalability and Optimization}
To manage the large volume of data and ensure smooth user interaction, \system uses several optimization strategies.

\paragraph{Search} For search functionalities, an index is built for each dataset with all metadata using Elasticsearch, allowing the backend to efficiently retrieve relevant conversations. To reduce the load during queries with a large number of matches, we employ two strategies: pagination, which retrieves results one page at a time with up to 30 conversations per page, and limiting the number of retrieved matches to 10,000 conversations per search.

\paragraph{Embedding Visualization - Frontend} Rendering a large number of conversation embeddings is computationally intensive for a browser, especially on mobile devices, and may lead to visual clutter with overlapping dots. To mitigate these issues, we use Deck.gl to render large numbers of points efficiently. Additionally, we restrict the visualization to a subset of 1,500 conversations per dataset, ensuring smooth rendering and clear visualization.

\paragraph{Embedding Visualization - Backend} On the backend, computing embeddings for a large number of conversations can introduce significant delays. To address this, we precompute the 2D coordinates for the subset of conversations selected for visualization. These precomputed results are then compressed using gzip and stored in a file, which is sent to the user upon their first visit to the embedding visualization page. The compressed file is approximately 1 MB in size and only needs to be downloaded once.

Although we only display a subset of conversations, users may still need to search the entire dataset. To support this, we integrate the embedding visualization with the Elasticsearch engine. When a user submits a query, we first search within the displayed subset of conversations (with an index built for this subset). If sufficient matches are found within the subset (with a default threshold of 100, adjustable up to 1,000), we simply highlight them and do not extend the search further. However, if there are not enough matches, we extend the search to the entire dataset using Elasticsearch, retrieve the relevant conversations (up to the threshold number), and embed and project them into 2D coordinates before sending them to the frontend for visualization. To speed up this process, we cache all computed coordinates in an SQLite database. Due to the need to dynamically compute coordinates for conversations not found in the cache, we chose parametric UMAP over t-SNE, as t-SNE does not learn a projection function, whereas parametric UMAP allows for quick projection of new conversations into lower-dimensional space.

\subsection{Performance Evaluation} To evaluate the efficiency of our system, we generated ten random keyword-based search queries and measured the execution time for each using our tool. On the filter-based search page, each query took an average of 0.47 seconds ($\pm0.06$s). In comparison, a naive for-loop-based approach using the HuggingFace Datasets library took 1148.89 seconds ($\pm25.28$s). For embedding visualization, the same measurement method was used, and each query took an average of 0.43 seconds ($\pm0.01$s).

\section{Use Cases}
This section presents several use cases that demonstrate the potential of \system. It is important to note that \system is designed primarily for exploratory data analysis rather than for final quantitative analysis.

\paragraph{Data} \system currently supports two datasets: WildChat~\citep{zhao2024wildchat} and LMSYS-Chat-1M~\citep{zheng2024lmsyschatm}. These datasets are integrated into the system by building Elasticsearch indices and precomputing the 2D coordinates of a randomly selected subset of conversations for embedding visualization.

\begin{figure*}[!t]
    \centering
    \begin{subfigure}[b]{0.49\linewidth}
        \centering
        \includegraphics[width=\linewidth]{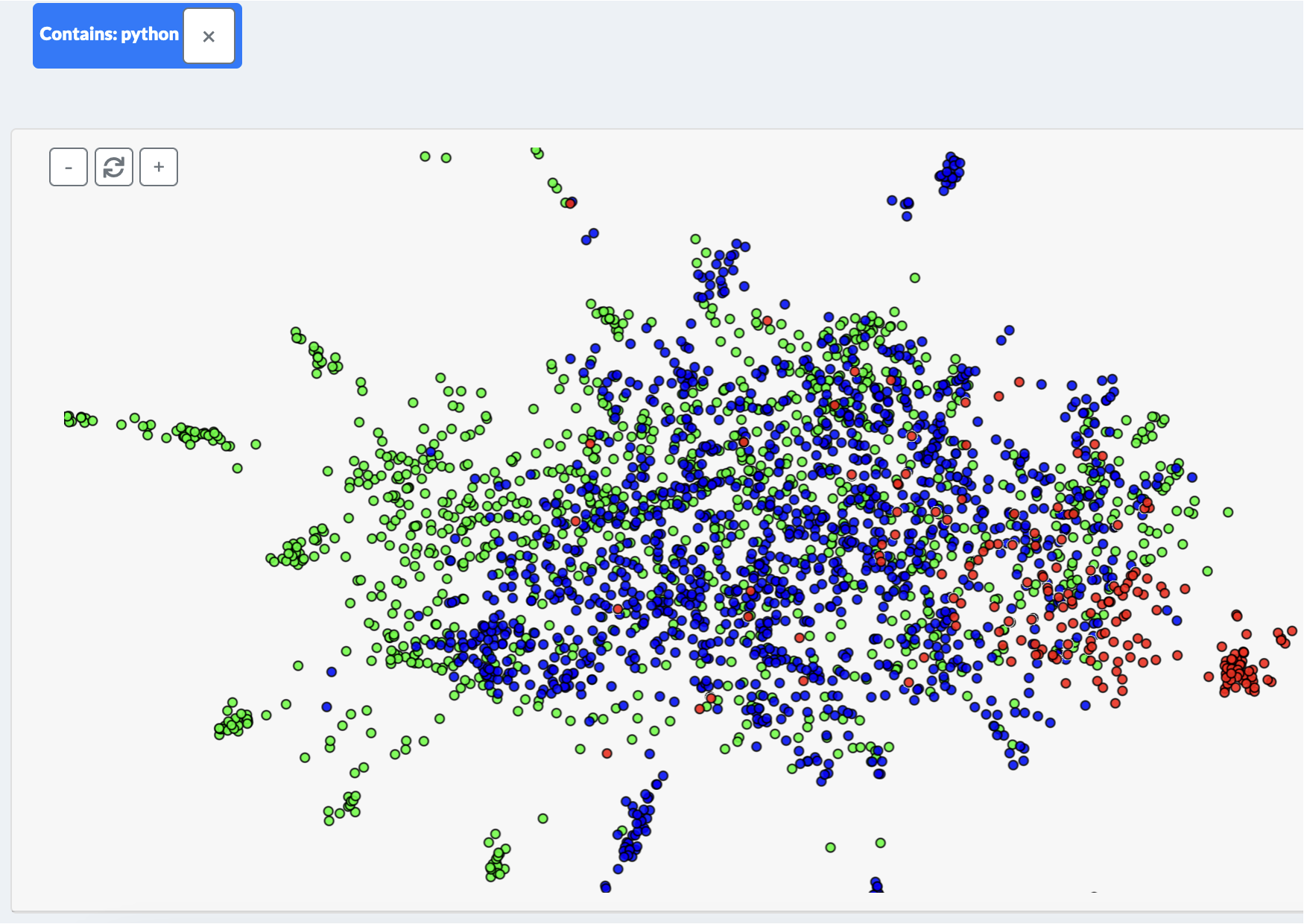}
        \caption{}
    \end{subfigure}
    \begin{subfigure}[b]{0.49\linewidth}
        \centering
        \includegraphics[width=\linewidth]{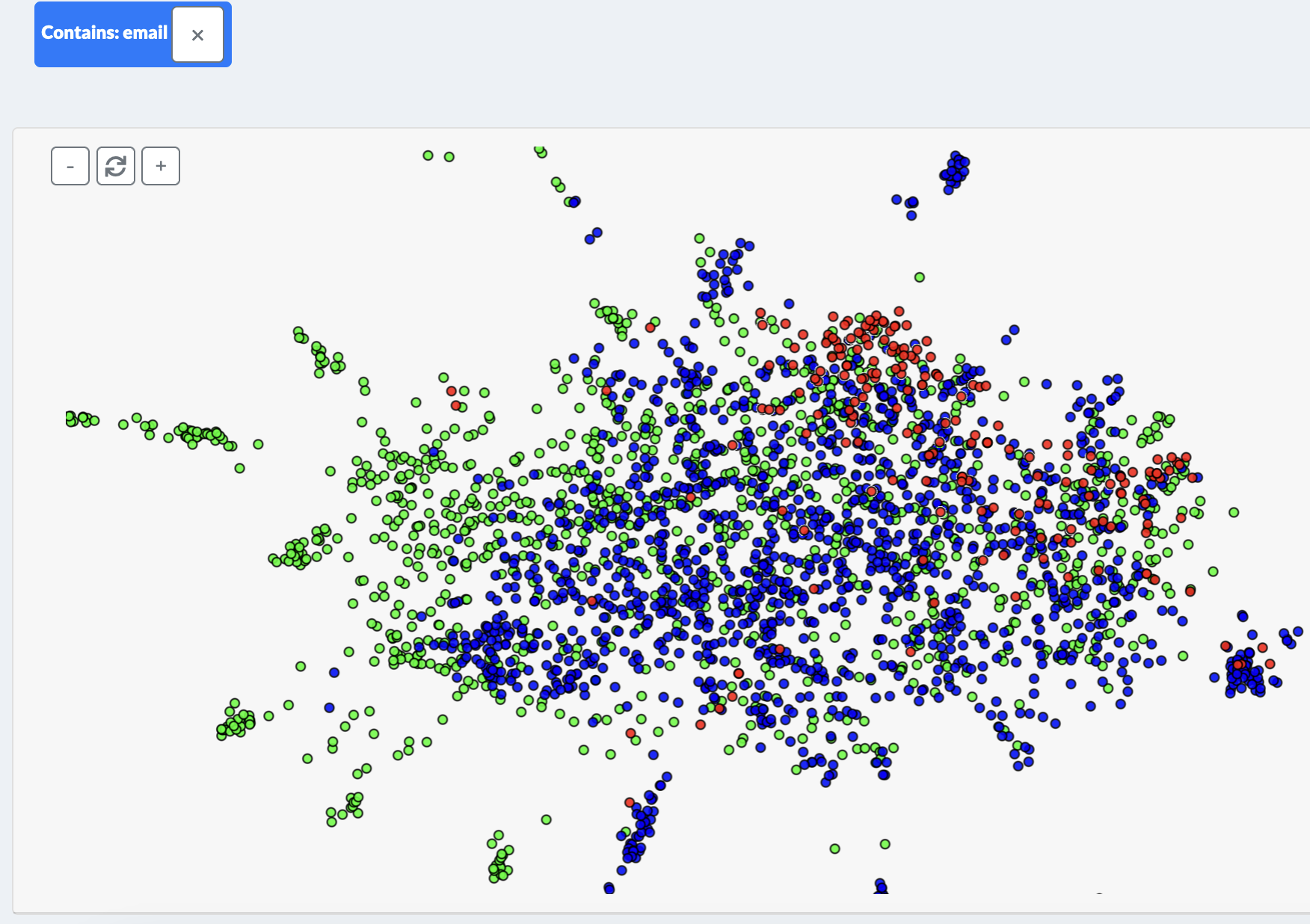}
        \caption{}
    \end{subfigure}

    \begin{subfigure}[b]{0.49\linewidth}
        \centering
        \includegraphics[width=\linewidth]{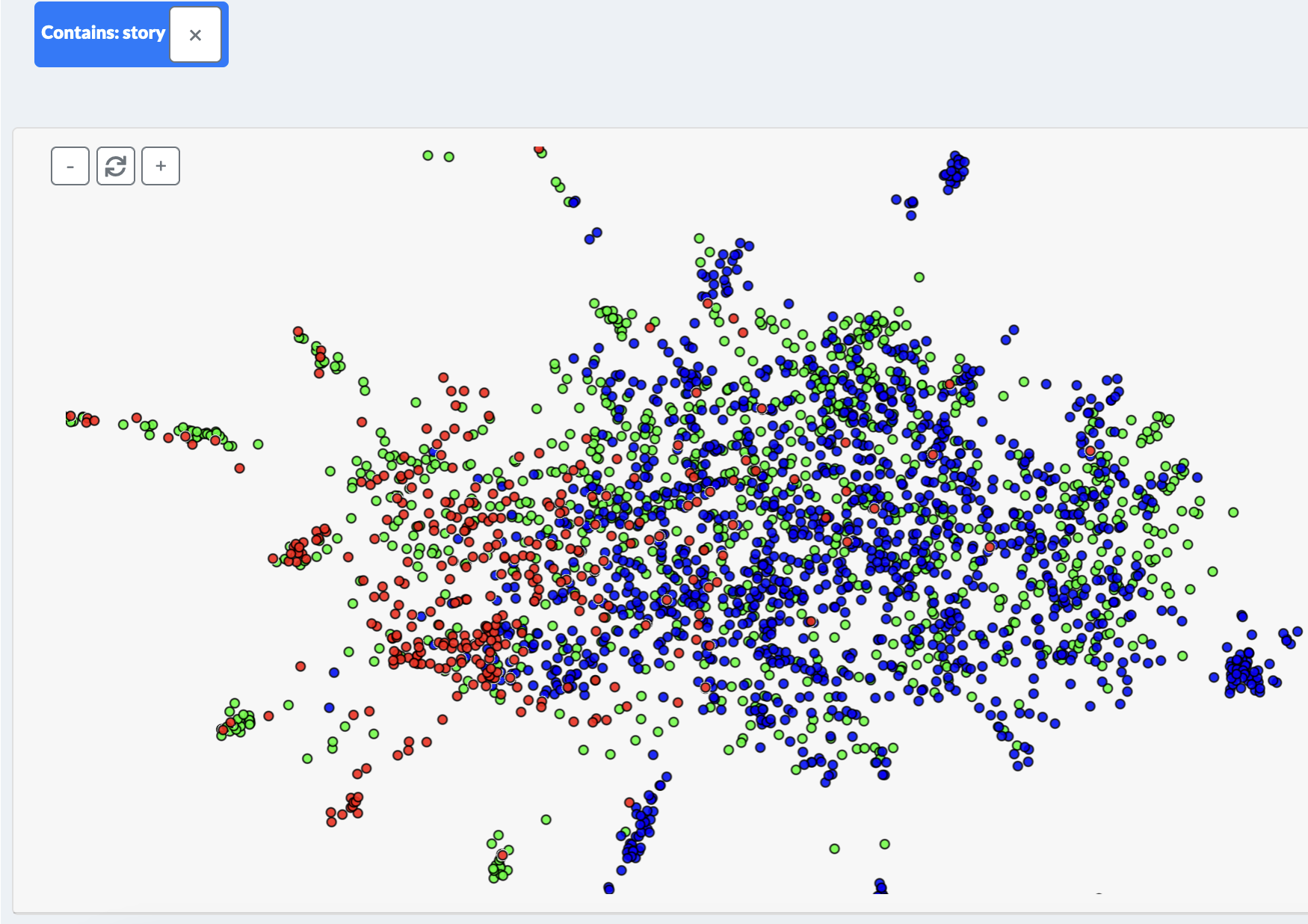}
        \caption{}
    \end{subfigure}
    \begin{subfigure}[b]{0.49\linewidth}
        \centering
        \includegraphics[width=\linewidth]{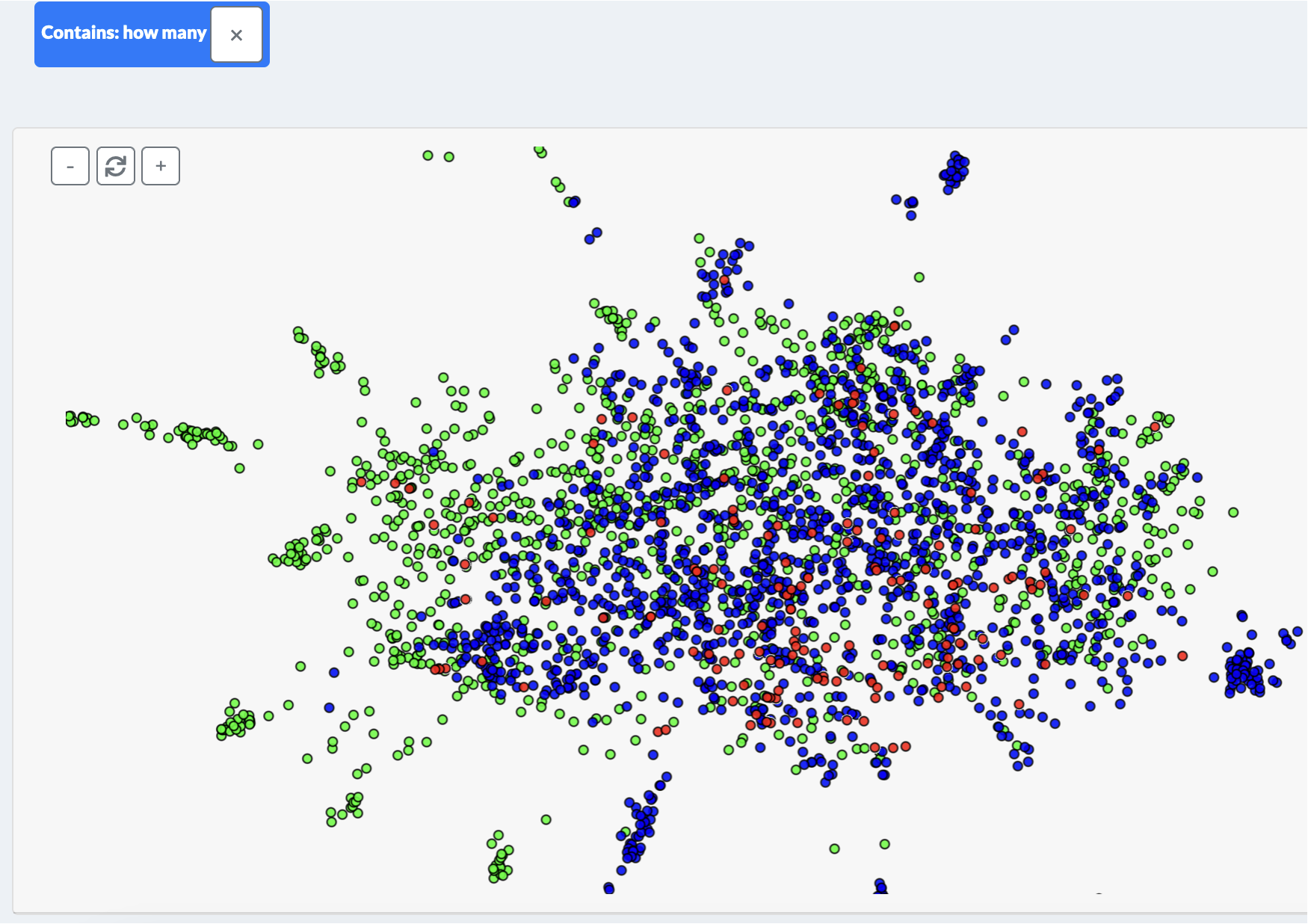}
        \caption{}
    \end{subfigure}
    \caption{\label{fig:vis}Major topic clusters.\protect\footnotemark{} (a) Coding (identified by searching for ``python''). (b) Writing assistance (identified by searching for ``email''). (c) Story generation (identified by searching for ``story''). (d) Math question answering (identified by searching for ``how many'').}
\end{figure*}

\subsection{Facilitating Chatbot Misuse Research}
One application of \system is in facilitating studies on chatbot misuse. We show here that \system is able to both reproduce existing studies on chatbot misuse and to discover new misuse cases.

\paragraph{Reproducing a Study on Journalist Misuse} In this use case, we replicate the findings of~\citet{brigham2024breakingnewscasestudies}, which identified instances of journalists misusing the chatbot behind WildChat to paraphrase existing articles for their work. To locate a specific instance mentioned in the study, we use the following quote from the original research:
\begin{quote}
write a new article out of the information in this article, do not make it obvious you are taking information from them but in very sensitive information give them credit.
\end{quote}

\footnotetext{These examples can be found at \url{https://wildvisualizer.com/embeddings/english?contains=python}, \url{https://wildvisualizer.com/embeddings/english?contains=email}, \url{https://wildvisualizer.com/embeddings/english?contains=story}, and \url{https://wildvisualizer.com/embeddings/english?contains=how\%20many}.}

To find this conversation, we enter the phrase \emph{you are taking information from them} in the ``Contains'' field on the search page and execute the search.\footnote{This case can be found at \url{https://wildvisualizer.com/?contains=you\%20are\%20taking\%20information\%20from\%20them}.} The search returns a single result, matching the case mentioned in the original paper. By clicking on the hashed IP address, we can view all conversations from this user, identifying all 15 conversations analyzed in the original study~\citep{brigham2024breakingnewscasestudies}.

\paragraph{Reproducing a Study on User Self-Disclosure} In another example, we replicate findings from a study on user self-disclosure behaviors by~\citet{mireshghallah2024trustbotdiscoveringpersonal}. We search for a key phrase from that paper: \emph{I have invited my father}.\footnote{This case can be found at \url{https://wildvisualizer.com/?contains=I\%20have\%20invited\%20my\%20father}.} Again, the search returns a single result, allowing us to find the conversation discussed in the study.

\paragraph{Discovering Additional Misuse Cases} \system also facilitates the discovery of additional misuse cases. For instance, by searching for conversations that contain both personally identifiable information (PII) and the term ``Visa Officer''\footnote{\url{https://wildvisualizer.com/?contains=Visa\%20Officer&redacted=true}}, we identified multiple entries from the same IP address. Further filtering based on this IP address revealed that the user appears to be affiliated with an immigration service firm and has disclosed sensitive client information.\footnote{\url{https://wildvisualizer.com/?hashed_ip=048b169ad0d18f2436572717f649bdeddac793967fb63ca6632a2f5dca14e4b8}}

\subsection{Visualizing and Comparing Topics} A powerful feature of the embedding visualization page in \system is its ability to visualize the overall distribution of topics, with conversations of similar topics positioned close to each other. In our previous discussion on embedding conversations, we illustrated language-specific clusters (\Cref{fig:embedding_language_clusters} in \Cref{appendix:embedding_language}). As another example, for English data, this visualization reveals that the embedding space can be roughly divided into four regions: coding (by searching for ``python''), writing assistance (by searching for ``email''), story generation (by searching for ``story''), and math question answering (by searching for ``how many''), as illustrated in \Cref{fig:vis}. This observation aligns with the findings in \citet{washpost_wildchat}.

This feature also allows for the comparison of topic distributions across different datasets. By inspecting regions with different colors, users can identify outliers, regions where one dataset is well-represented while the other is not, and areas where both datasets overlap. By hovering over these regions, patterns in the types of conversations can be observed. For example, we found that WildChat contains more conversations related to creating writing and an outlier cluster of Midjourney prompt generation (see \Cref{fig:midjourney}) compared to LMSYS-Chat-1M, while LMSYS-Chat-1M has outlier clusters of conversations about chemistry (see \Cref{fig:chemistry}).

\subsection{Characterizing User-Specific Patterns}
\system can also be used to visualize the topics of all conversations associated with a specific user on the embedding map. For example, \Cref{fig:userspecific} displays all conversations of a single user, revealing two main topic clusters: coding-related and email writing-related.



\section{Related Work}

\paragraph{HuggingFace Dataset Viewer} HuggingFace's Dataset Viewer~\citep{lhoest-etal-2021-datasets}\footnote{\url{https://huggingface.co/docs/dataset-viewer/en/index}} provides basic search functionalities for datasets hosted on HuggingFace. However, it is designed for general dataset visualization and is not specifically tailored for conversational datasets. For example, while it offers useful statistics, navigating JSON-formatted conversations in a table format can be cumbersome and lacks the intuitive visualization needed for exploring conversational data.

\paragraph{Paper Visualization Tools} The ACM Fellows' Citation Visualization tool\footnote{\url{https://mojtabaa4.github.io/acm-citations/}} embeds ACM Fellows based on their contribution statements. While its interface shares many similarities with the embedding visualization page of \system, it focuses on publication data rather than conversational data. Another relevant work is \citet{yen2024scholarly}, which visualizes papers in a similar manner, with an added conversational component that allows users to interact with the visualizations by asking questions. However, it is also primarily designed for academic papers rather than large-scale chat datasets.

\paragraph{Browser Tools for Chat Visualization} Several browser-based tools exist for chat visualization, such as ShareGPT\footnote{\url{https://sharegpt.com}}, which allows users to share their conversations. However, ShareGPT lacks support for searching large-scale chat datasets. Similarly, browser extensions like ShareLM\footnote{\url{https://chromewebstore.google.com/detail/nldoebkdaiidhceaphmipeclmlcbljmh}} enable users to upload and view their conversations, and ChatGPT History Search\footnote{\url{https://chatgpthistorysearch.com/en}} offers search functionality for a user's personal conversations. However, these tools are not designed for the exploration or analysis of large-scale chat datasets.

\paragraph{Large-scale Data Analysis Tools}
Specialized tools like ConvoKit~\citep{chang-etal-2020-convokit} provide a framework for analyzing dialogue data. In comparison, \system is designed to offer an intuitive interface for interactively exploring and visualizing chat datasets. This makes \system particularly useful for preliminary data exploration and hypothesis generation. Another notable tool, WIMBD~\citep{elazar2024whats}, supports the analysis and comparison of large text corpora, offering functionalities such as searching for documents containing specific queries and counting statistics like n-gram occurrences. Although WIMBD can handle larger datasets, \system offers additional features, such as embedding visualization, providing a more comprehensive toolkit for chat dataset exploration.


\section{Conclusion}
In this paper, we introduced \system, an interactive web-based tool designed for exploring large-scale conversational datasets. By combining powerful search functionalities with intuitive visualization capabilities, \system enables researchers to uncover patterns and gain insights from vast collections of user-chatbot interactions. The system's scalability optimizations ensure efficient handling of million-scale datasets, while maintaining a responsive and user-friendly experience.

\system fills a gap in existing tools by providing a specialized platform for visualizing and exploring chat datasets, which are inherently challenging to analyze using generic dataset viewers. Our use cases demonstrate the tool's potential to replicate and extend existing research on chatbot misuse and user self-disclosure, as well as to facilitate topic-based conversation exploration.


\section*{Acknowledgments}
This work is supported by ONR grant N00014-24-1-2207, NSF grant DMS-2134012, and an NSERC Discovery grant. We also thank Bing Yan, Pengyu Nie, and Jiawei Zhou for their valuable feedback.

\bibliography{anthology, anthology2, custom}

\begin{thebibliography}{14}
\providecommand{\natexlab}[1]{#1}

\bibitem[{Brigham et~al.(2024)Brigham, Gao, Kohno, Roesner, and Mireshghallah}]{brigham2024breakingnewscasestudies}
Natalie~Grace Brigham, Chongjiu Gao, Tadayoshi Kohno, Franziska Roesner, and Niloofar Mireshghallah. 2024.
\newblock \href {https://arxiv.org/abs/2406.13706} {Breaking news: Case studies of generative ai's use in journalism}.
\newblock \emph{Preprint}, arXiv:2406.13706.

\bibitem[{Chang et~al.(2020)Chang, Chiam, Fu, Wang, Zhang, and Danescu-Niculescu-Mizil}]{chang-etal-2020-convokit}
Jonathan~P. Chang, Caleb Chiam, Liye Fu, Andrew Wang, Justine Zhang, and Cristian Danescu-Niculescu-Mizil. 2020.
\newblock \href {https://doi.org/10.18653/v1/2020.sigdial-1.8} {{C}onvo{K}it: A toolkit for the analysis of conversations}.
\newblock In \emph{Proceedings of the 21th Annual Meeting of the Special Interest Group on Discourse and Dialogue}, pages 57--60, 1st virtual meeting. Association for Computational Linguistics.

\bibitem[{Elazar et~al.(2024)Elazar, Bhagia, Magnusson, Ravichander, Schwenk, Suhr, Walsh, Groeneveld, Soldaini, Singh, Hajishirzi, Smith, and Dodge}]{elazar2024whats}
Yanai Elazar, Akshita Bhagia, Ian~Helgi Magnusson, Abhilasha Ravichander, Dustin Schwenk, Alane Suhr, Evan~Pete Walsh, Dirk Groeneveld, Luca Soldaini, Sameer Singh, Hannaneh Hajishirzi, Noah~A. Smith, and Jesse Dodge. 2024.
\newblock \href {https://openreview.net/forum?id=RvfPnOkPV4} {What's in my big data?}
\newblock In \emph{The Twelfth International Conference on Learning Representations}.

\bibitem[{Lhoest et~al.(2021)Lhoest, Villanova~del Moral, Jernite, Thakur, von Platen, Patil, Chaumond, Drame, Plu, Tunstall, Davison, {\v{S}}a{\v{s}}ko, Chhablani, Malik, Brandeis, Le~Scao, Sanh, Xu, Patry, McMillan-Major, Schmid, Gugger, Delangue, Matussi{\`e}re, Debut, Bekman, Cistac, Goehringer, Mustar, Lagunas, Rush, and Wolf}]{lhoest-etal-2021-datasets}
Quentin Lhoest, Albert Villanova~del Moral, Yacine Jernite, Abhishek Thakur, Patrick von Platen, Suraj Patil, Julien Chaumond, Mariama Drame, Julien Plu, Lewis Tunstall, Joe Davison, Mario {\v{S}}a{\v{s}}ko, Gunjan Chhablani, Bhavitvya Malik, Simon Brandeis, Teven Le~Scao, Victor Sanh, Canwen Xu, Nicolas Patry, Angelina McMillan-Major, Philipp Schmid, Sylvain Gugger, Cl{\'e}ment Delangue, Th{\'e}o Matussi{\`e}re, Lysandre Debut, Stas Bekman, Pierric Cistac, Thibault Goehringer, Victor Mustar, Fran{\c{c}}ois Lagunas, Alexander Rush, and Thomas Wolf. 2021.
\newblock \href {https://doi.org/10.18653/v1/2021.emnlp-demo.21} {Datasets: A community library for natural language processing}.
\newblock In \emph{Proceedings of the 2021 Conference on Empirical Methods in Natural Language Processing: System Demonstrations}, pages 175--184, Online and Punta Cana, Dominican Republic. Association for Computational Linguistics.

\bibitem[{Malik(2023)}]{malik2023chatgpt}
Aisha Malik. 2023.
\newblock \href {https://techcrunch.com/2023/11/06/openais-chatgpt-now-has-100-million-weekly-active-users/} {{OpenAI’s ChatGPT now has 100 million weekly active users}}.
\newblock Accessed: 2024-08-04.

\bibitem[{McInnes et~al.(2020)McInnes, Healy, and Melville}]{mcinnes2020umapuniformmanifoldapproximation}
Leland McInnes, John Healy, and James Melville. 2020.
\newblock \href {https://arxiv.org/abs/1802.03426} {Umap: Uniform manifold approximation and projection for dimension reduction}.
\newblock \emph{Preprint}, arXiv:1802.03426.

\bibitem[{Merrill and Lerman(2024)}]{washpost_wildchat}
Jeremy~B. Merrill and Rachel Lerman. 2024.
\newblock \href {https://www.washingtonpost.com/technology/2024/08/04/chatgpt-use-real-ai-chatbot-conversations/} {What do people really ask chatbots? it’s a lot of sex and homework.}
\newblock \emph{The Washington Post}.
\newblock Accessed: 2024-08-27.

\bibitem[{Mireshghallah et~al.(2024)Mireshghallah, Antoniak, More, Choi, and Farnadi}]{mireshghallah2024trustbotdiscoveringpersonal}
Niloofar Mireshghallah, Maria Antoniak, Yash More, Yejin Choi, and Golnoosh Farnadi. 2024.
\newblock \href {https://arxiv.org/abs/2407.11438} {Trust no bot: Discovering personal disclosures in human-llm conversations in the wild}.
\newblock \emph{Preprint}, arXiv:2407.11438.

\bibitem[{Rush and Strobelt(2020)}]{rush2020miniconfvirtualconference}
Alexander~M. Rush and Hendrik Strobelt. 2020.
\newblock \href {https://arxiv.org/abs/2007.12238} {Miniconf -- a virtual conference framework}.
\newblock \emph{Preprint}, arXiv:2007.12238.

\bibitem[{Sainburg et~al.(2021)Sainburg, McInnes, and Gentner}]{sainburg2021parametricumapembeddingsrepresentation}
Tim Sainburg, Leland McInnes, and Timothy~Q Gentner. 2021.
\newblock \href {https://arxiv.org/abs/2009.12981} {Parametric umap embeddings for representation and semi-supervised learning}.
\newblock \emph{Preprint}, arXiv:2009.12981.

\bibitem[{van~der Maaten and Hinton(2008)}]{JMLR:v9:vandermaaten08a}
Laurens van~der Maaten and Geoffrey Hinton. 2008.
\newblock \href {http://jmlr.org/papers/v9/vandermaaten08a.html} {Visualizing data using t-sne}.
\newblock \emph{Journal of Machine Learning Research}, 9(86):2579--2605.

\bibitem[{Yen et~al.(2024)Yen, Brus, Yan, Lin, and Zhao}]{yen2024scholarly}
Ryan Yen, Yelizaveta Brus, Leyi Yan, Jimmy Lin, and Jian Zhao. 2024.
\newblock \href {https://ryanyen2.github.io/papers/scholet.pdf} {Scholarly exploration via conversations with scholars-papers embedding}.

\bibitem[{Zhao et~al.(2024)Zhao, Ren, Hessel, Cardie, Choi, and Deng}]{zhao2024wildchat}
Wenting Zhao, Xiang Ren, Jack Hessel, Claire Cardie, Yejin Choi, and Yuntian Deng. 2024.
\newblock \href {https://openreview.net/forum?id=Bl8u7ZRlbM} {Wildchat: 1m chat{GPT} interaction logs in the wild}.
\newblock In \emph{The Twelfth International Conference on Learning Representations}.

\bibitem[{Zheng et~al.(2024)Zheng, Chiang, Sheng, Li, Zhuang, Wu, Zhuang, Li, Lin, Xing, Gonzalez, Stoica, and Zhang}]{zheng2024lmsyschatm}
Lianmin Zheng, Wei-Lin Chiang, Ying Sheng, Tianle Li, Siyuan Zhuang, Zhanghao Wu, Yonghao Zhuang, Zhuohan Li, Zi~Lin, Eric Xing, Joseph~E. Gonzalez, Ion Stoica, and Hao Zhang. 2024.
\newblock \href {https://openreview.net/forum?id=BOfDKxfwt0} {{LMSYS}-chat-1m: A large-scale real-world {LLM} conversation dataset}.
\newblock In \emph{The Twelfth International Conference on Learning Representations}.

\end{thebibliography}

\clearpage
\appendix

\section{\label{appendix:embedding_page_mobile}Embedding Visualization on Mobile Devices}
\begin{figure}[!t]
    \centering
    \includegraphics[width=\linewidth]{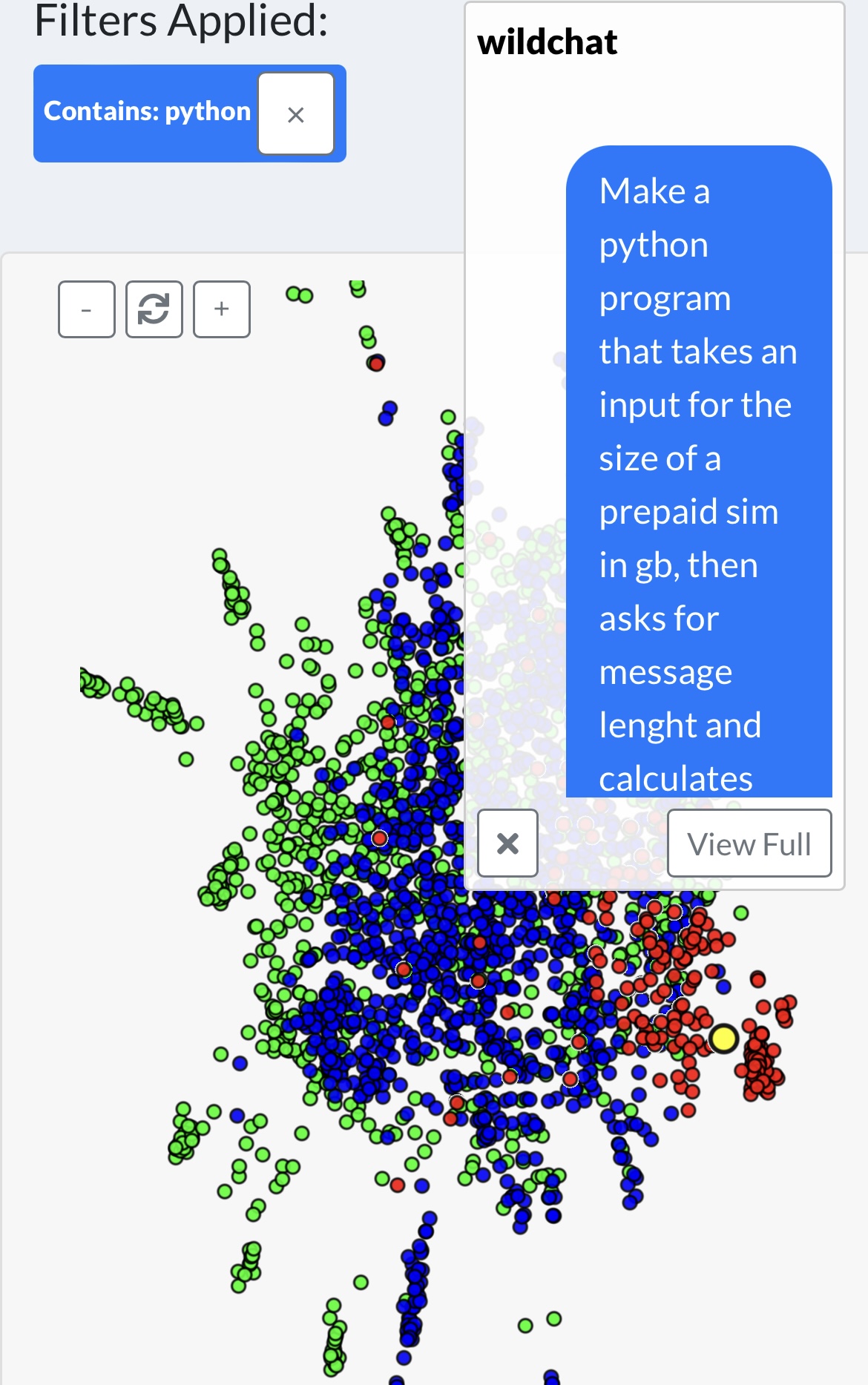}
    \caption{\label{fig:embedding_page_mobile}\system Embedding Visualization on Mobile Devices. Tapping a dot displays a preview with options to view the full conversation or close the preview. This example can be viewed at \url{https://wildvisualizer.com/embeddings/english?contains=python} on a mobile device.}
\end{figure}

\Cref{fig:embedding_page_mobile} shows a screenshot of the embedding visualization page on mobile devices. Since mobile devices do not support hover interactions, we adapted the interface by using a tap gesture for displaying previews. Additionally, a button is provided to view the full conversation, replacing the click action used on desktop devices.

\section{\label{appendix:embedding_language}Language-Specific Clusters}
\begin{figure*}[!t]
    \centering
    \begin{subfigure}[b]{0.75\linewidth}
        \centering
        \includegraphics[width=\linewidth]{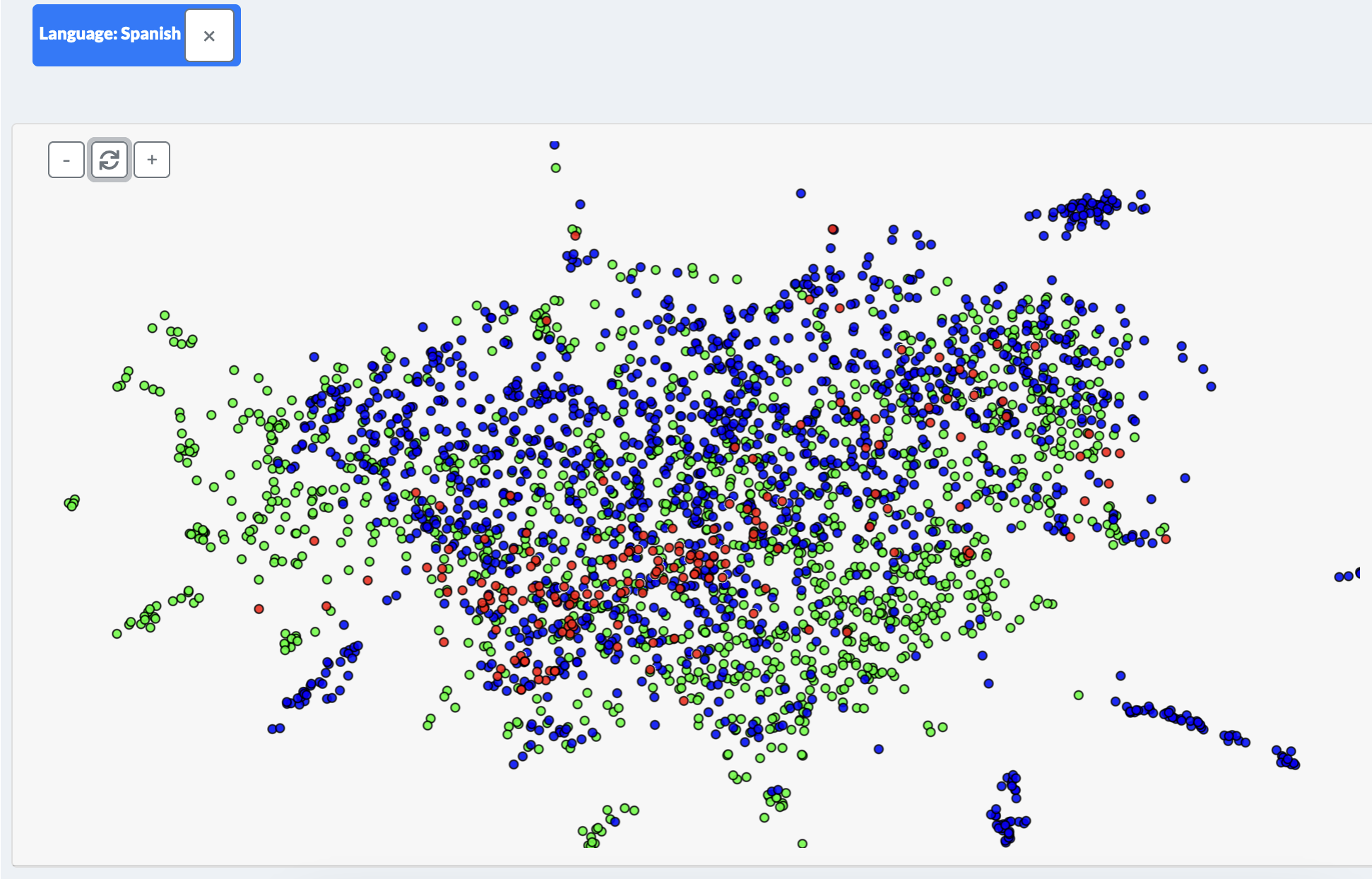}
    \end{subfigure}
    
    \begin{subfigure}[b]{0.75\linewidth}
        \centering
        \includegraphics[width=\linewidth]{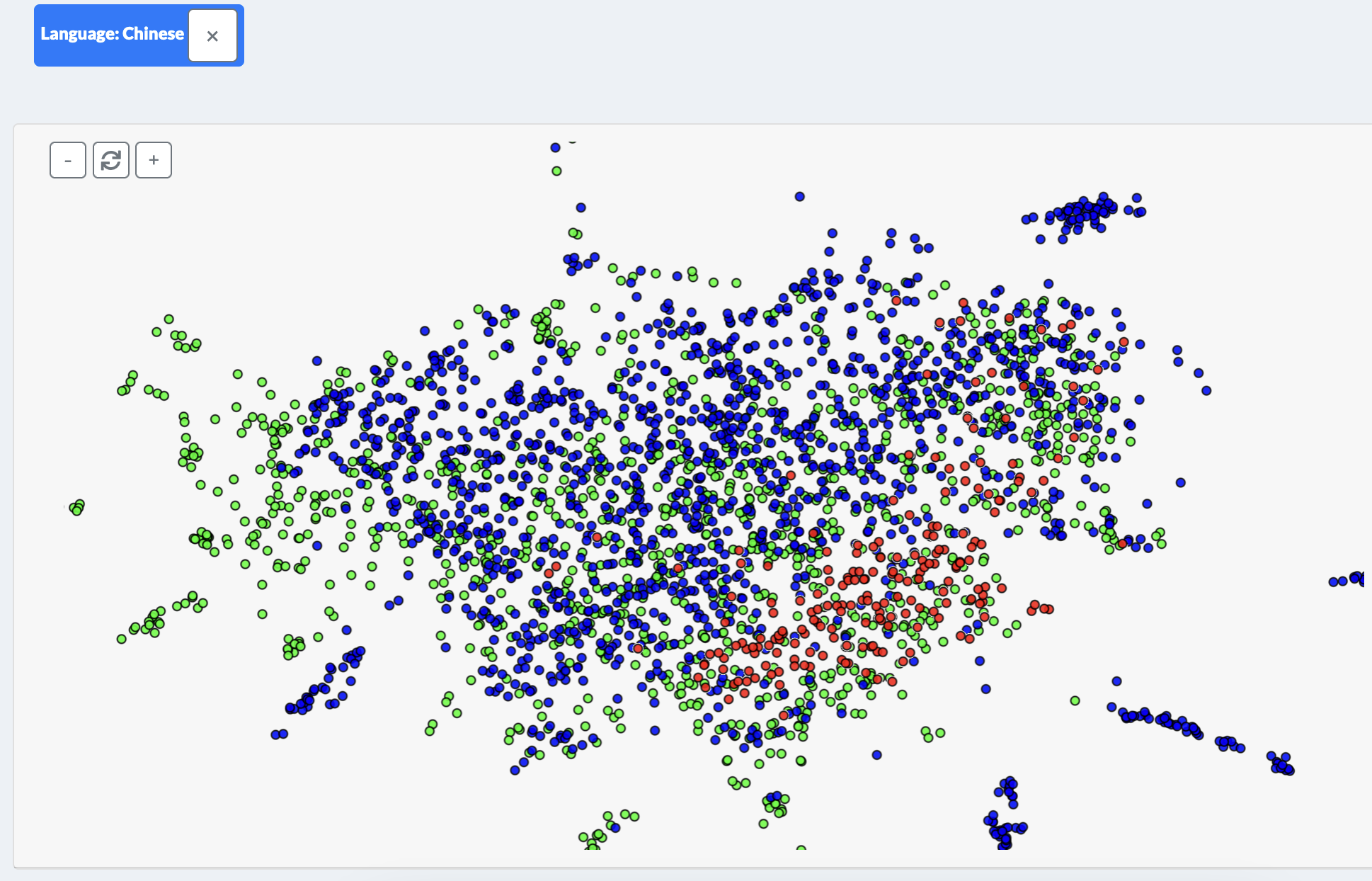}
    \end{subfigure}

     \begin{subfigure}[b]{0.75\linewidth}
        \centering
        \includegraphics[width=\linewidth]{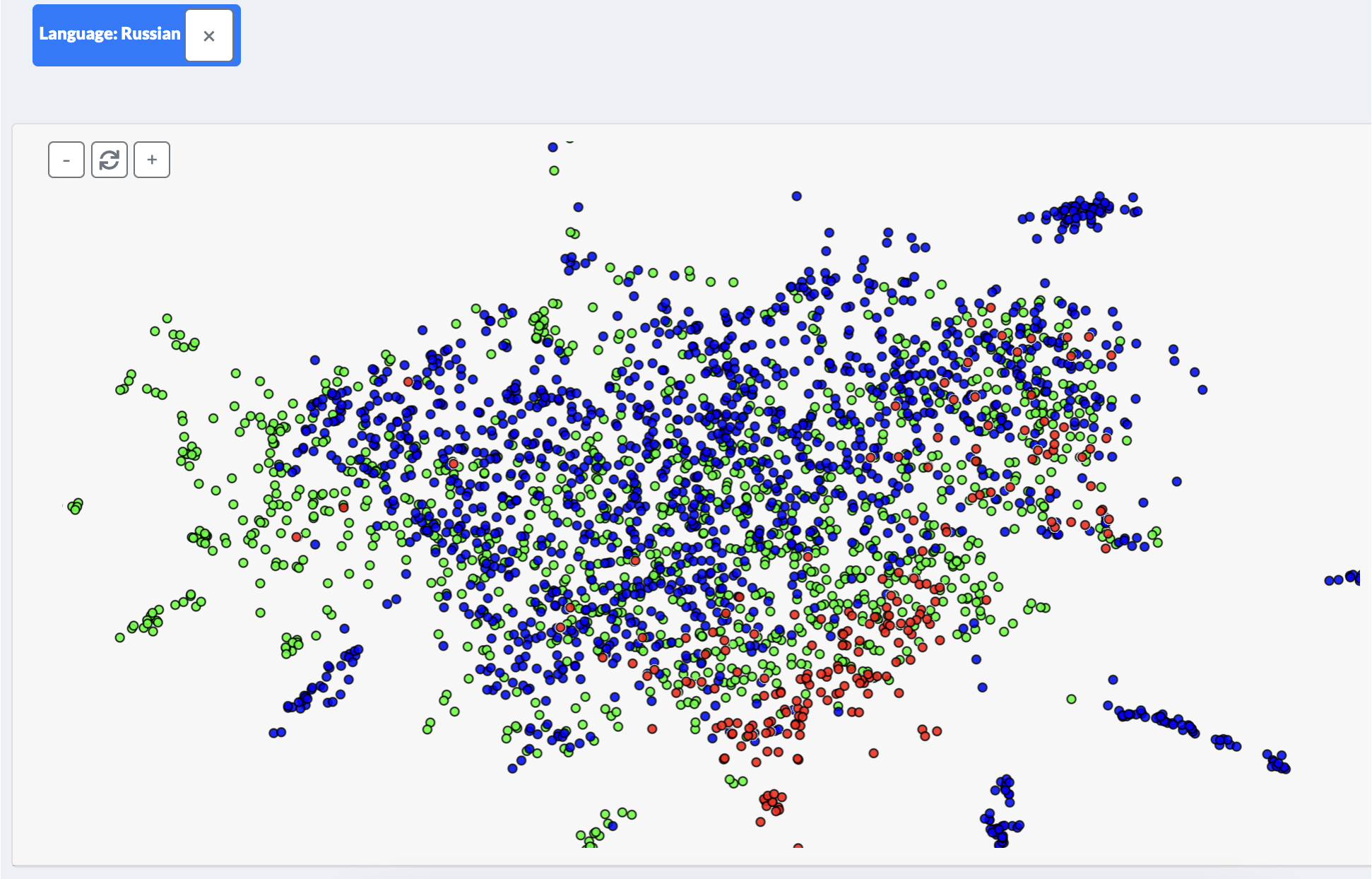}
    \end{subfigure}
    \caption{\label{fig:embedding_language_clusters}Language-specific clusters. Top: Spanish. Middle: Chinese. Bottom: Russian. These can be found at \url{https://wildvisualizer.com/embeddings?language=Spanish}, \url{https://wildvisualizer.com/embeddings?language=Chinese}, and \url{https://wildvisualizer.com/embeddings?language=Russian}.}
\end{figure*}

When visualizing all conversations together on the embedding visualization page, clusters based on language emerge, such as the Spanish, Chinese, and Russian clusters in \Cref{fig:embedding_language_clusters}.

\section{\label{appendix:embedding_switch}Switching Embedding Visualization Language}
\begin{figure*}[!t]
    \centering
    \includegraphics[width=\linewidth]{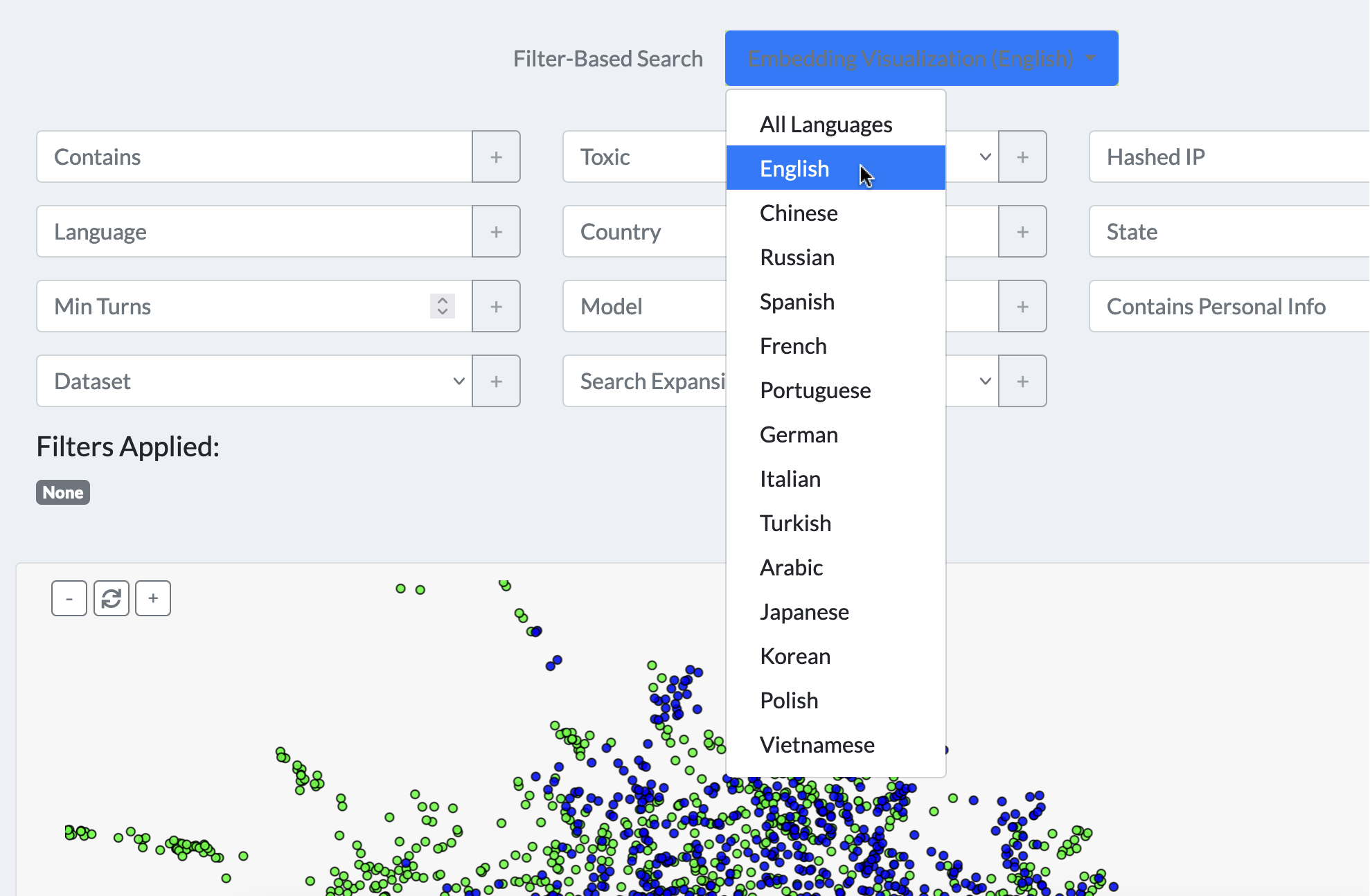}
    \caption{\label{fig:embedding_switch}Switching the embedding visualization language. This will load conversations in the selected language and apply the corresponding trained parametric UMAP projection model to embed conversations. This example is available at \url{https://wildvisualizer.com/embeddings/english}.}
\end{figure*}

\Cref{fig:embedding_switch} shows a screenshot of switching the embedding visualization language. This will load a subset of conversations in the selected language only and utilize the corresponding trained parametric UMAP model to embed conversations.

\section{\label{sec:appendix_A}Conversation Details Page}
\begin{figure*}[!htp]
    \centering
    \includegraphics[width=\linewidth]{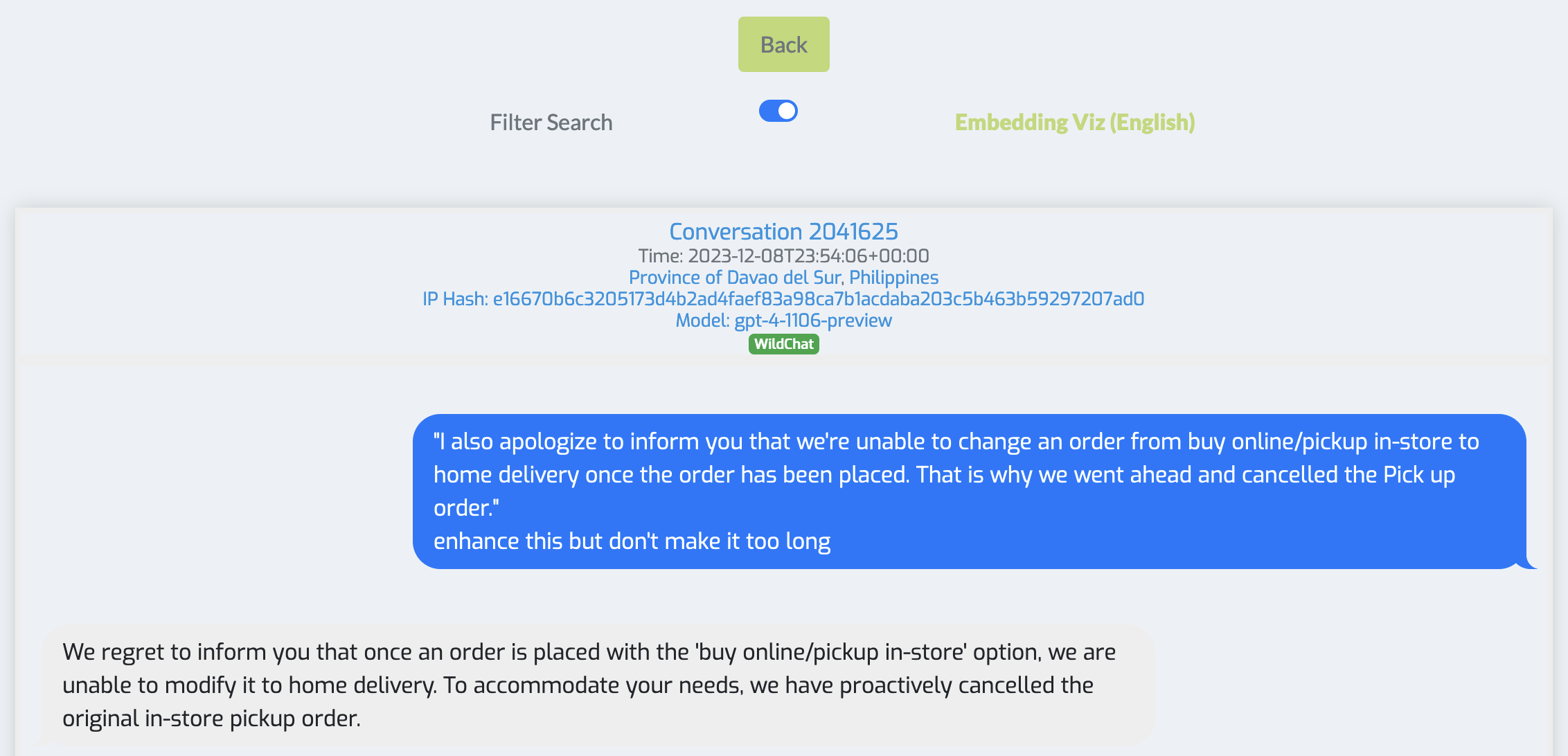}
    \caption{\label{fig:conversation_page}\system Conversation Details Page. This page provides a detailed view of individual conversations, displaying all interactions between the user and the chatbot. Key metadata, including the conversation ID, timestamp, geographic location, and the model used, are presented at the top. Clicking any metadata field filters based on its value, redirecting users to either the filter-based search page or the embedding visualization page, depending on the original navigation path. A toggle switch at the top allows users to control this behavior. This example can be found at \url{https://wildvisualizer.com/conversation/wildchat/2041625?from=embedding&lang=english}.}
\end{figure*}

\Cref{fig:conversation_page} shows a screenshot of the conversation details page, where all metadata fields are displayed alongside the dialogue content. Clicking any metadata field will filter the conversations based on the selected value. Depending on how the user navigated to this page---either from the filter-based search page or the embedding visualization page---the filtering action will redirect the user back to the respective page. A toggle switch at the top allows users to control this behavior.


\section{Visualizing and Comparing Topic Distributions}
\begin{figure*}[!htp]
    \centering
    \includegraphics[width=\linewidth]{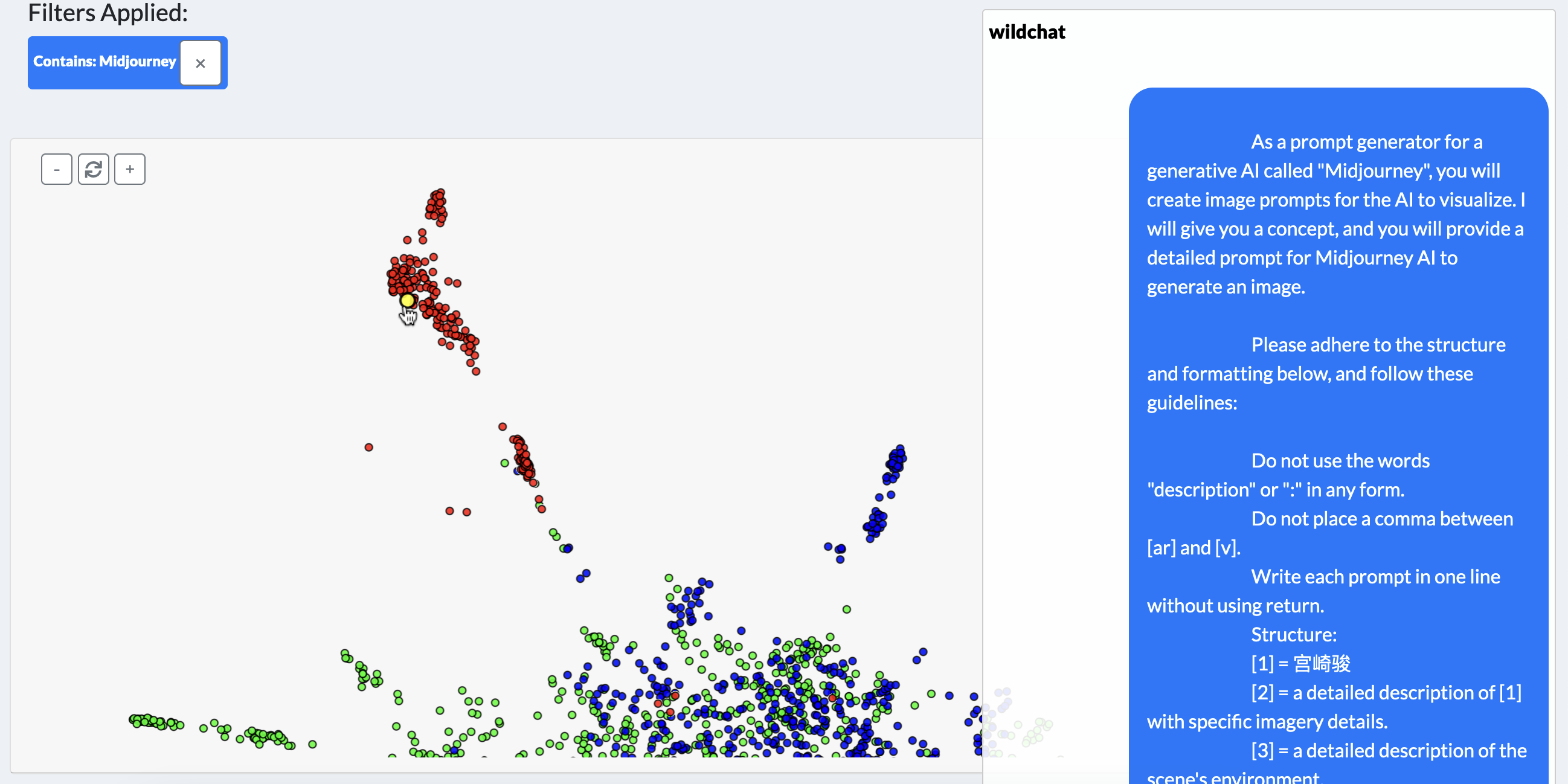}
    \caption{\label{fig:midjourney}Embedding visualization showing an outlier cluster related to Midjourney prompt engineering in WildChat. This example can be found at \url{https://wildvisualizer.com/embeddings/english?contains=Midjourney}.}
\end{figure*}

\begin{figure*}[!htp]
    \centering
    \includegraphics[width=\linewidth]{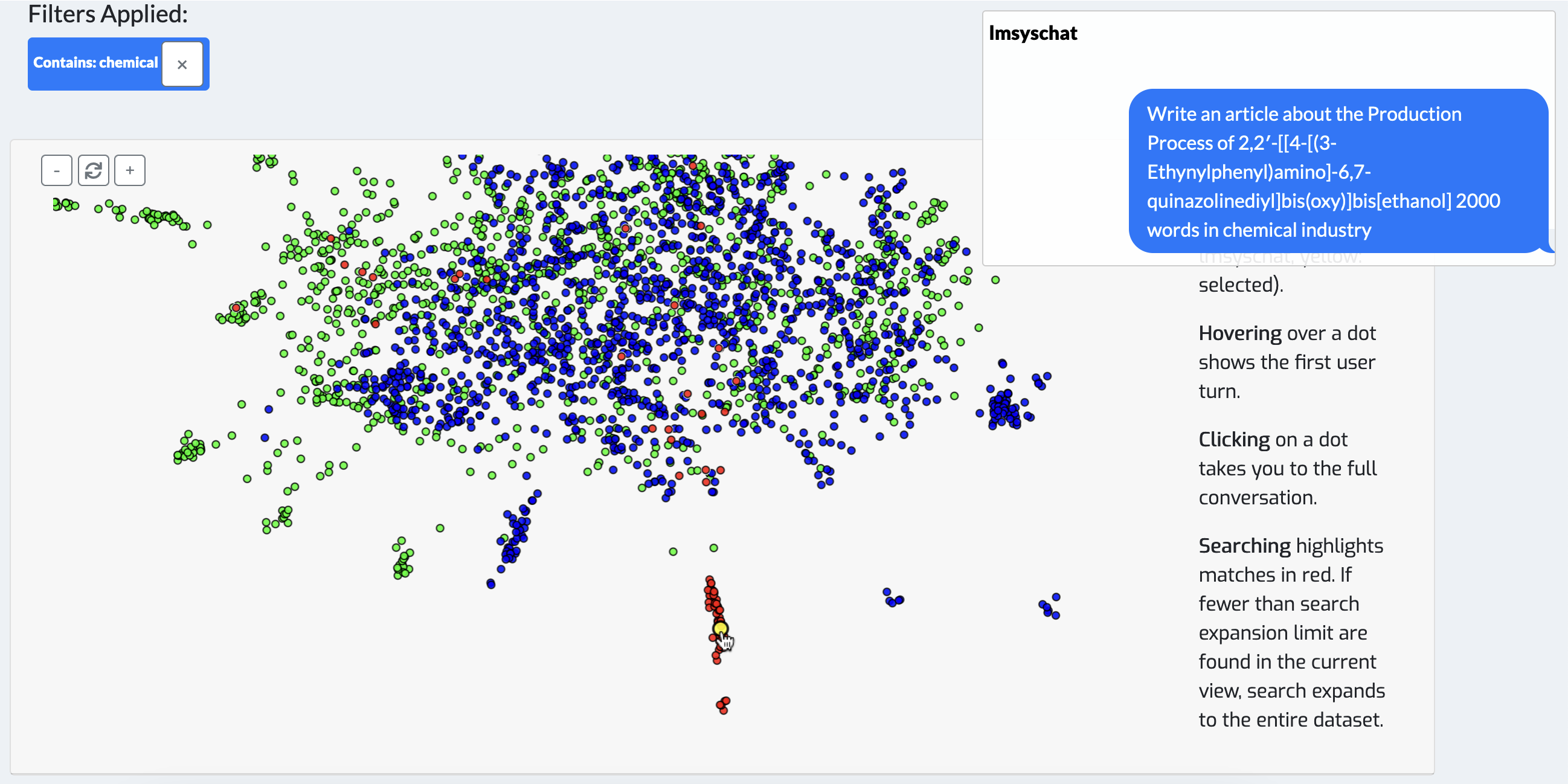}
    \caption{\label{fig:chemistry}Embedding visualization showing an outlier cluster related to chemistry questions in LMSYS-Chat-1M. This example can be found at \url{https://wildvisualizer.com/embeddings/english?contains=chemical}.}
    \vspace{2.2in}
\end{figure*}

The embedding visualization highlights distinct outlier clusters in the dataset. One notable cluster in the WildChat dataset involves Midjourney prompt engineering, where users ask the chatbot to generate detailed prompts for use with Midjourney, as shown in \Cref{fig:midjourney} (this phenomenon was also noted by \citet{washpost_wildchat}). Another distinct outlier cluster comprises chemistry-related questions in LMSYS-Chat-1M, illustrated in \Cref{fig:chemistry}.\footnote{\href{https://future-xy.github.io/}{Yao Fu} discovered this phenomenon and shared it with the authors.}

\begin{figure*}[t!]
    \centering
    \includegraphics[width=\linewidth]{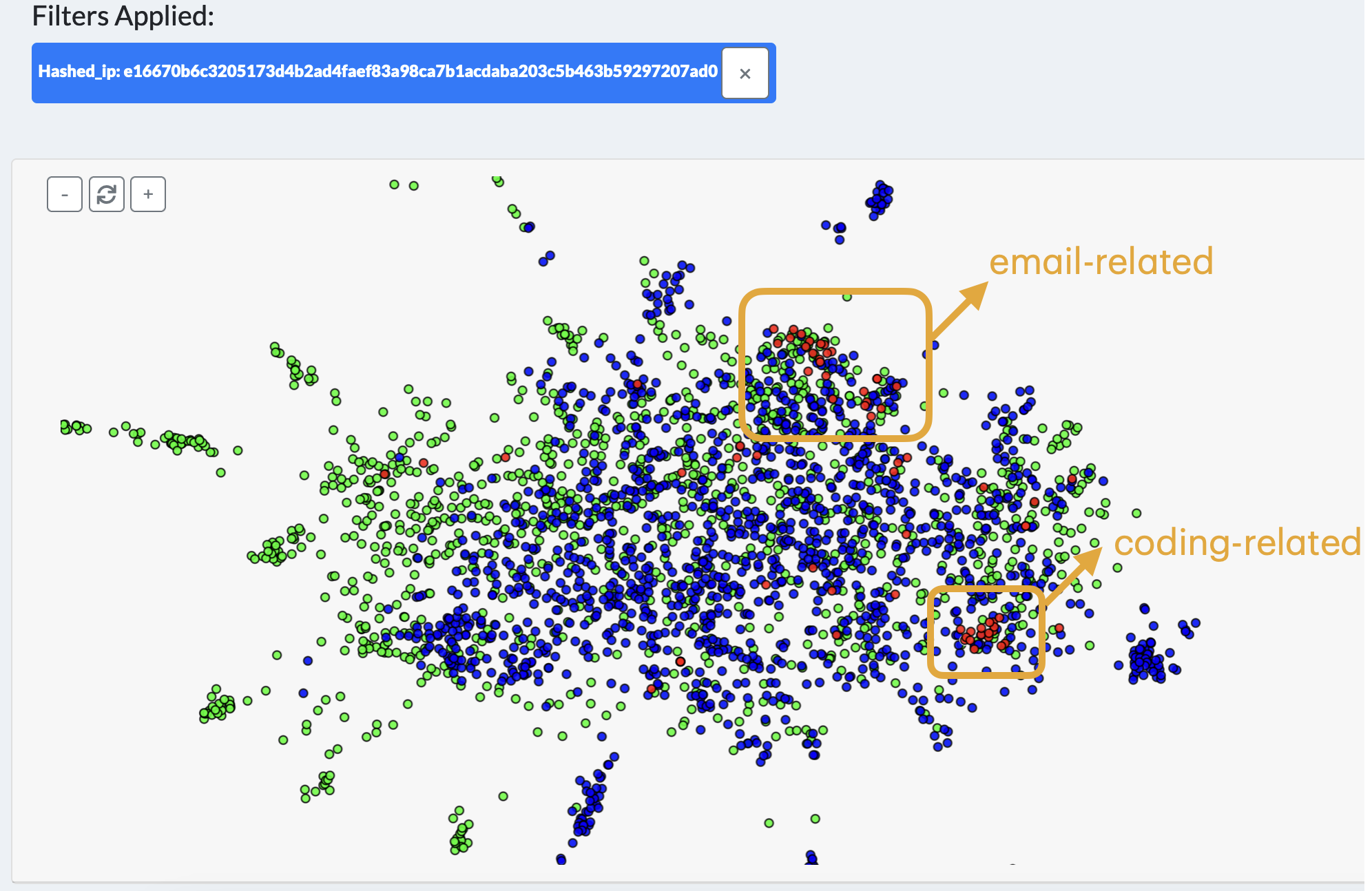}
    \caption{\label{fig:userspecific}Embedding visualization of all conversations from a single user. Two major clusters are evident: one related to coding and the other to email writing assistance. This example can be found at \url{https://wildvisualizer.com/embeddings/english?hashed_ip=e16670b6c3205173d4b2ad4faef83a98ca7b1acdaba203c5b463b59297207ad0}.}
    \vspace{4.85in}
\end{figure*}

\section{Characterizing User-Specific Patterns}
\system can be used to visualize the topics of all conversations associated with a specific user on the embedding map. For example, \Cref{fig:userspecific} displays all conversations from a single user, revealing two main topic clusters: coding-related and email writing-related.

\end{document}